\documentclass[11pt]{article}

\PassOptionsToPackage{table}{xcolor}
\usepackage[preprint]{acl}

\usepackage{times}
\usepackage{latexsym}

\usepackage[T1]{fontenc}

\usepackage[utf8]{inputenc}

\usepackage{microtype}

\usepackage{inconsolata}

\usepackage{graphicx}

\usepackage{amsmath,amssymb,amsthm}
\newtheorem{definition}{Definition}
\usepackage{algorithm}
\usepackage{algpseudocode}
\usepackage{booktabs}
\usepackage{enumitem}
\usepackage{multirow}
\usepackage{rotating}

\newcommand{\first}[1]{\textbf{#1}}  
\newcommand{\second}[1]{\underline{#1}}  
\newcommand{\gain}[1]{\textcolor{blue}{{\small $\uparrow$#1}}}  
\newcommand{\drop}[1]{\textcolor{red}{{\small $\downarrow$#1}}}  

\newcommand{\concept}[1]{\textnormal{\textsc{#1}}}

\title{Spatial-Agent: Agentic Geo-spatial Reasoning with Scientific Core Concepts}

\author{
  \textbf{Riyang Bao\textsuperscript{1*}},
  \textbf{Cheng Yang\textsuperscript{2*}},
  \textbf{Dazhou Yu\textsuperscript{1}},
  \textbf{Zhexiang Tang\textsuperscript{2}},
  \textbf{Gengchen Mai\textsuperscript{3}},
  \textbf{Liang Zhao\textsuperscript{1}}
  \\[0.5em]
  \textsuperscript{1}Emory University,
  \textsuperscript{2}Rutgers University,
  \textsuperscript{3}University of Texas at Austin
  \\
  \small{\textsuperscript{*}Equal contribution. \quad \textbf{Correspondence:} Liang Zhao}
}

\begin{document}
\maketitle
\begin{abstract}
Geospatial reasoning is essential for real-world applications such as urban analytics, transportation planning, and disaster response. However, existing LLM-based agents often fail at genuine geospatial computation, relying instead on web search or pattern matching while hallucinating spatial relationships. We present \textbf{Spatial-Agent}, an AI agent grounded in foundational theories of spatial information science. Our approach formalizes geo-analytical question answering as a \textit{concept transformation} problem, where natural-language questions are parsed into executable workflows represented as \textit{GeoFlow Graphs}---directed acyclic graphs with nodes corresponding to spatial concepts and edges representing transformations. Drawing on spatial information theory, Spatial-Agent extracts spatial concepts, assigns functional roles with principled ordering constraints, and composes transformation sequences through template-based generation. Extensive experiments on MapEval-API and MapQA benchmarks demonstrate that Spatial-Agent significantly outperforms existing baselines including ReAct and Reflexion, while producing interpretable and executable geospatial workflows.
\end{abstract}

\section{Introduction}
Geospatial reasoning is essential for numerous real-world applications, including urban analytics, transportation planning, environmental monitoring, disaster response, and public health \citep{li2021neural,li2023location,mai2021geographic,mai2025towards}. As geospatial datasets and modern GIS platforms 
continue to proliferate, users increasingly expect natural-language interfaces capable of handling complex geo-analytical questions \cite{yu2025spatial,scheider2020ontology,scheider2021geo}. However, despite the impressive capabilities of contemporary large language models (LLMs), we observe that current agent-style systems, including function-calling LLMs and commercial AI agents, do not perform genuine geospatial reasoning. 
\begin{figure}[t]
    \centering
    \includegraphics[width=\columnwidth]{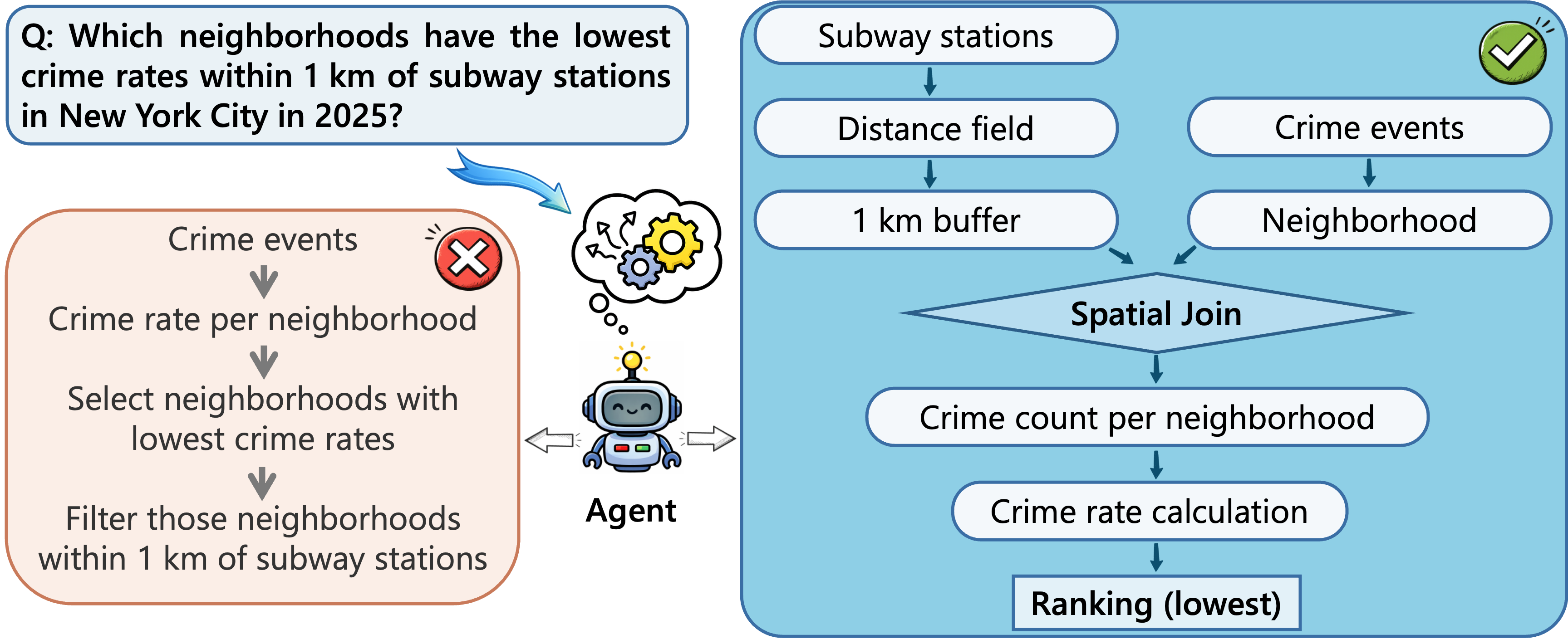}
    \caption{LLM-intuitive but incorrect workflow (left) vs. correct concept transformation (right). The incorrect applies spatial constraints after aggregation; the correct computes crime rates within spatial context first.}
    \label{fig:intro}
\end{figure}

Prior work shows that LLM-based agents lack inherent spatial awareness, relying on web search or textual pattern matching rather than computational spatial analysis~\citep{yan2023inherent,zhang2025geoanalystbench}. As illustrated in Figure~\ref{fig:intro}, they hallucinate spatial relationships, fail on geometric or topological predicates, and cannot construct valid workflows~\citep{wang2025geohallucination,ji2025foundation}.

This reflects a deeper limitation: geo-analytical questions require
procedural, multi-step reasoning over spatial data, which fundamentally differs from declarative QA. Classical GIScience research has long emphasized that geographical phenomena are inherently computational rather than purely linguistic \cite{goodchild1992geographical, miller2015data, goodchild2007towards}. Answering a geo-analytical question often involves: (i) identifying core spatial entities (objects, events, fields,
networks), (ii) selecting appropriate spatial operators (buffering, overlay,
routing, aggregation), (iii) ordering operators into an executable workflow, and (iv) grounding abstract instructions into concrete GIS tools (e.g., PostGIS, ArcGIS, QGIS). These operations are geometric, topological, and sometimes spatiotemporal in nature, and thus beyond the representational power of language-only models.

A promising direction in GIScience suggests that geo-analytical questions encode implicit procedural knowledge. Kuhn's theory of
\textit{core concepts} \cite{kuhn2012core} and subsequent work by Scheider et al. \cite{scheider2020ontology} identifies foundational building blocks of spatial information, such as \textit{Object}, \textit{Field}, \textit{Event} and \textit{Network}. Complementing these, GeoAnQu research \cite{xu2023grammar} highlights \textit{functional roles} (\textit{Measure}, \textit{Condition}, \textit{Subcondition}, \textit{Support}, \textit{Extent}, etc.) that encode the procedural structure of geo-analytical questions. By combining core concepts and functional roles, GeoAnQu demonstrates that natural-language questions can be mapped into concept transformations, forming the basis of GIS workflows represented as directed acyclic graphs (DAGs). Yet this framework remains rule-based, offline, and detached from modern AI agent architectures.

In this paper, we ask: \textbf{What would it take for an AI agent to truly understand and execute geo-analytical questions?} Despite advances in tool-augmented LLMs, current agents remain limited in the geospatial domain. They often misinterpret spatial entities and relations, lack a procedural understanding of how geo-analytical tasks unfold, and cannot reliably map natural-language questions to coherent sequences of spatial operations. These systems also struggle to manage intermediate spatial states or ground their reasoning in computational GIS tools, which leads to answers that are largely descriptive rather than operational. As a result, existing agents fail to produce verifiable and executable geospatial analyses.

To address these challenges, we present \textbf{Spatial-Agent}, a geospatial AI agent grounded in foundational theories of spatial information. We formalize geo-analytical question answering as a \textit{concept transformation} problem, where natural-language questions are parsed into executable workflows represented as \textit{GeoFlow Graphs}. Drawing on core concepts and functional roles, Spatial-Agent establishes a principled intermediate representation that bridges language and computation. Specifically, the agent (i) extracts spatial concepts from questions and instantiates them as graph nodes, (ii) identifies functional roles that impose ordering constraints, (iii) composes transformation edges through a template-based approach that leverages recurring geo-analytical patterns, and (iv) executes the workflow via tool invocations, grounding its final response in verifiable computational results rather than parametric knowledge alone.

Our contributions are summarized as follows:
\begin{itemize}
    \item We present Spatial-Agent, which enables geospatial reasoning by uncovering the implicit structure of spatial questions and generating coherent, executable workflows.
    \item We propose a compositional GeoFlow Graph generation approach based on macro-templates, capturing recurring geo-analytical patterns and improving structural validity via template matching and IO-port composition.
    \item Extensive evaluations show that SpatialAgent delivers significantly better correctness, interpretability, and executable workflow generation than existing agent baselines, bridging the gap between natural-language reasoning and computational GIS.
\end{itemize}

\section{Related Work}

\paragraph{Geospatial Question Answering.}
Geospatial question answering \cite{mai2021geographic} is a sub-domain of question answering focusing on questions that involve geographic entities, concepts, and/or require geospatial computation. 
Early geospatial QA systems focused on factoid-style questions over knowledge graphs \cite{mai2020se}. GeoQA \cite{punjani2018template} and its successor GeoQA2 \cite{kefalidis2024question} answer questions over DBpedia and YAGO2geo using template-based SPARQL query generation. The GeoQuestions1089 benchmark \cite{kefalidis2023benchmarking} provides 1,089 questions with GeoSPARQL queries for evaluation. TourismQA \cite{contractor2021answering,contractor2021joint} was constructed as a tourism-oriented geospatial QA dataset focusing on retrieving points of interest (POIs) from spatial databases based on text-to-SQL approaches according to specified geospatial and semantics constraints. 
More recently, MapQA \cite{li2025mapqa} was introduced as a similar text-to-SQL style POI retrieval benchmark with diverse geospatial constraits. 
However, these systems primarily handle declarative queries rather than procedural, multi-step geo-analytical reasoning. \citet{scheider2021geo} and its follow-up work, GeoAnQu \cite{xu2023grammar}, move toward geo-analytical questions by identifying functional roles and spatial concept transformations, but remain rule-based and offline. 

\paragraph{LLM-based Agents and Tool Use.}
Recent advances in LLM agents have demonstrated impressive capabilities in tool-augmented reasoning. ReAct \cite{yao2022react} introduced the thought-action-observation loop for interleaved reasoning and acting. Toolformer \cite{schick2023toolformer} enables self-supervised tool use learning, while CodeAct \cite{wang2024executable} shows that generating executable code outperforms JSON-based actions by up to 20\%. In the geospatial domain, LLM-Geo \cite{li2023autonomous} and GeoGPT \cite{zhang2024geogpt} demonstrate autonomous GIS capabilities using GPT-4 for geoprocessing workflow generation and code execution. GeoAgent \cite{chen2024llm} integrates RAG with Monte Carlo Tree Search for geospatial data processing, and GTChain \cite{zhang2025geospatial} fine-tunes LLaMA for geospatial tool-use chains. 

\paragraph{Spatial Core Concepts and Workflow Composition.}
Kuhn's theory of core concepts \cite{kuhn2012core} provides foundational building blocks for spatial information, including \textit{Object}, \textit{Field}, \textit{Event}, and \textit{Network}. \citet{scheider2020ontology} formalized these into the Core Concept Data Type (CCD) ontology, enabling semantic constraints for GIS workflow automation. \citet{kruiger2021loose} demonstrated that loose programming with CCD types can automatically construct valid workflows for tasks like accessibility assessment and spatial interpolation. In parallel, neural program synthesis \cite{devlin2017robustfill, zhong2023hierarchical} has shown success in generating structured programs from specifications. 

\section{Spatial-Agents}

\begin{figure*}[t]
    \centering
    \includegraphics[width=1\textwidth]{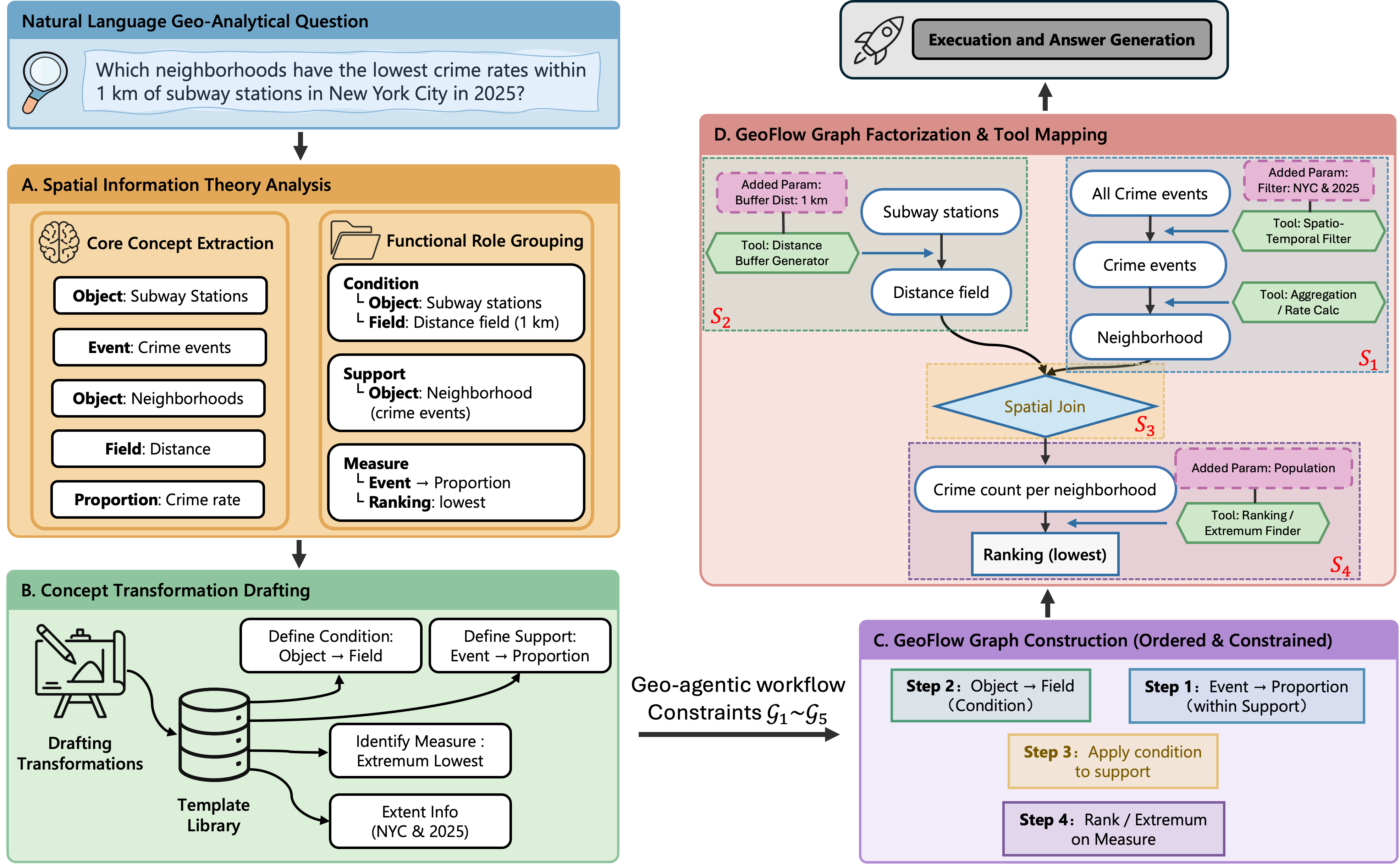}
    \caption{Overview of Spatial-Agent: (A) Spatial information theory analysis extracts core concepts and assigns functional roles; (B) Concept transformation drafting composes templates from the library; (C) GeoFlow Graph construction produces an ordered and constrained graph; (D) Graph factorization maps to executable tools for execution and answer generation.}
    \label{fig:framework}
\end{figure*}

\subsection{Problem formulation of Agentic Geo-Spatial Reasoning}
Unlike existing LLM-based agents that rely on general-purpose planning and reasoning, Spatial-Agent is designed around a key insight: geo-analytical questions require grounding natural-language semantics into computational spatial representations. This poses three fundamental challenges that distinguish geo-analytical reasoning from general-purpose tasks:
\begin{itemize}
\item \textbf{Semantic Domain vs.\ Spatial Domain.}
LLM agents represent spatial relationships as linguistic patterns rather than geometric structures. For instance, ``airport'' and ``terminal'' are linguistically interchangeable, yet spatially the terminal is contained within the airport---a distinction critical for distance computation but invisible to language models.

\item \textbf{Cognitive Reasoning vs.\ Spatial Reasoning.}
LLMs treat geospatial tools as black-box APIs without understanding the procedural structure of geo-analytical questions. They lack a principled way to decompose spatial queries into ordered operations, often producing invalid workflows.

\item \textbf{Spatial Orchestration vs.\ Spatial Execution.}
Even when LLMs understand individual spatial operations, they struggle to orchestrate them into executable workflows where intermediate states must be tracked and data dependencies respected.
\end{itemize}

To address these challenges, Spatial-Agent introduces: (1) explicit spatial grounding that transforms semantic descriptions into computational representations with precise geometric meanings; (2) a theory-driven framework leveraging GIScience core concepts and functional roles to decompose geo-analytical questions; and (3) the GeoFlow Graph as an intermediate representation bridging spatial knowledge with agent execution.

\paragraph{System Overview.}
As illustrated in Figure~\ref{fig:framework}, Spatial-Agent operates through a multi-stage pipeline: (1) \textit{Spatial Information Theory Analysis} (\S\ref{sec:grounding}) extracts core concepts from the question and groups them by functional roles; (2) \textit{Concept Transformation Drafting} retrieves templates to define transformation patterns between concepts; (3) \textit{GeoFlow Graph Construction} (\S\ref{sec:orchestration}) assembles transformations into an ordered graph following role-based precedence constraints; (4) \textit{GeoFlow Graph Factorization \& Tool Mapping} converts the graph into an executable form with concrete operators; and (5) the agent executes the workflow and generates a grounded response.
\subsection{Geospatial Concept Grounding}
\label{sec:grounding}
To ground semantic concepts in spatial analysis, we must map abstract, non-spatial concepts from their original domains to concrete geo-objects in the spatial domain, and subsequently derive the corresponding geographic relations among them. This grounding process requires two complementary formalisms: \textit{core concepts} that define what spatial entities exist, and \textit{functional roles} that specify how they participate in analysis. GIScience has long recognized that such grounding follows principled patterns governed by geospatial semantics \cite{goodchild1992geographical,kuhn2012core,mai2021geographic,scheider2021geo}.

To formalize this grounding process, we represent the result as a \textit{GeoFlow Graph}. Given a natural-language geo-analytical question $q$, our goal is to derive a GeoFlow Graph $G$ that represents an executable geospatial workflow:
\begin{equation}
f: q \mapsto G
\end{equation}
where $G = (V, E, \lambda, \rho)$ consists of concept nodes $V$, transformation edges $E \subseteq V \times V$, a concept labeling function $\lambda: V \rightarrow \mathcal{C}$ mapping nodes to spatial concepts (defined below based on spatial core concept theory), and a role assignment function $\rho: V \rightarrow \mathcal{R}$ assigning functional roles (defined below based on geo-analytical reasoning principles). Each directed edge $(v_i, v_j) \in E$ represents a \textit{transformation}---a semantic change from one spatial concept to another (e.g., from a place name to coordinates via geocoding). Each transformation is realized by an \textit{operator} $\omega \in \Omega$, a concrete computational function that takes input concepts and produces output concepts.

Here, we define the spatial concept space $\mathcal{C}$ as a set of \textit{core concepts} that ground semantic representations into geographic primitives~\cite{kuhn2012core}:
\begin{equation}
\mathcal{C} = \left\{\begin{array}{l}
\concept{Location}, \concept{Object}, \concept{Field}, \concept{Event}, \\
\concept{Network}, \concept{Amount}, \concept{Proportion}
\end{array}\right\}
\end{equation}
These seven primitives, widely recognized in geographic information theory, capture the fundamental building blocks of geographic phenomena and provide a vocabulary for representing spatial entities extracted from natural-language questions (see Appendix~\ref{app:core-concepts} for detailed definitions and examples).

Crucially, we operationalize these abstract concepts into a mathematical framework that LLMs can manipulate. Each concept $c \in \mathcal{C}$ is not merely a label but carries associated type signatures, valid transformations, and composability rules. This formalization transforms the descriptive spatial theory into a computational substrate: the agent can query concept compatibility, validate transformation sequences, and reason about workflow correctness through symbolic operations on $\mathcal{C}$.

While core concepts define \textit{what} spatial entities are, we also need to specify \textit{how} they participate in geo-analytical reasoning. To this end, we introduce \emph{functional roles} to explicitly encode the procedural structure of geo-analytical questions ~\cite{xu2023grammar}:
\begin{equation}
\mathcal{R} = \left\{\begin{aligned}
  &\concept{Extent}, \concept{TExtent}, \concept{SubCond},\\
  &\concept{Cond}, \concept{Support}, \concept{Measure}
\end{aligned}\right\}.
\end{equation}

We also distinguish between \emph{contextual roles}
(\textsc{Extent}, \textsc{TExtent}) and \emph{procedural roles}
(\textsc{SubCond}, \textsc{Cond}, \textsc{Support}, \textsc{Measure}).
Contextual roles constrain data collection but do not participate in the procedural ordering of transformations (e.g., ``New York City'' as \concept{Extent} and ``2025'' as \concept{TExtent} in Figure~\ref{fig:framework}   (B)). For procedural roles, we define a precedence relation
\(\prec\) derived from the execution semantics of geo-analytical workflows: $\concept{SubCond} \prec \concept{Cond} \prec \concept{Support} \prec \concept{Measure}$.

This ordering reflects the inherent structure of geo-analytical workflows: sub-conditions restrict candidate entities, conditions constrain supports, supports establish the spatial basis, and measurements are the final outputs. By making this procedural structure explicit, the agent follows transformation orders derived from geographic first principles rather than hallucinating arbitrary operation sequences (see Appendix~\ref{app:functional-roles} for detailed role definitions).

\subsection{Spatial Orchestration}
\label{sec:orchestration}
With the concept space $\mathcal{C}$ and functional roles $\mathcal{R}$ established, we describe how to construct GeoFlow Graph $G$ from a question. The challenge is determining which transformation edges to include: an edge $(v_i, v_j) \in E$ should be added when the resolution of $v_j$ depends on the output of $v_i$.

$V$ includes both \textit{explicit} concepts directly mentioned in the question and \textit{implicit} concepts inferred to complete the workflow. For example, given ``What is the driving time from my hotel to the nearest coffee shop?'', the explicit concepts are \textsc{Hotel}, \textsc{CoffeeShop}, and \textsc{DrivingTime}. However, computing driving time requires a \textsc{RoadNetwork} (prerequisite data source) and an intermediate \textsc{Route} connecting the two locations---neither of which is explicitly mentioned. The generation process (\S\ref{sec:dag-gen}) is responsible for identifying and instantiating these implicit nodes.

\paragraph{Geoflow graph factorization.}
GeoFlow Graphs capture concept-level transformations, where nodes represent spatial core concepts and edges denote semantic dependencies. However, these edges do not directly correspond to executable operators in an agentic system: many geo-analytical transformations jointly consume multiple input concepts, while the original graph encodes them as separate edges. To make GeoFlow Graphs executable, we reinterpret them as a factorized operator–concept hypergraph $G'=(V',E')$ such that there is a bijective mapping between them $G \leftrightarrow G'$, where $V'$ are the nodes consisting of spatial core concept nodes and factor nodes. The concept nodes (e.g., the round nodes such as ``Subway stations'' and ``Crime events'' in Figure~\ref{fig:framework} (D)) are directly extracted from the question, while factor nodes (e.g., the pink boxes such as ``Buffer Dist: 1 km'' and ``Population'') represent supplementary parameters required for operator execution. Each operator takes concept nodes along with supplementary parameters as inputs and produces one or more downstream concept nodes as outputs.
This factorization enables many-to-many relationships that traditional sequential graphs cannot directly express. The operators span geocoding, spatial search, routing, geometric computation, and trip optimization (see Appendix~\ref{app:operators} for details).

\paragraph{Geo-agentic workflow Constraints.}
Based on fundamental GIScience principles \cite{tobler1970computer,janowicz2012geospatial,janowicz2012observation,janowicz2022know}, we formalize the structural regularities of geographic phenomena as explicit constraints. A valid GeoFlow Graph must satisfy:

(1)~\textbf{Acyclicity} ($\mathcal{G}_1$): $\mathcal{G}_1 = \{G \mid \nexists \text{ cycle in } E\}$, ensuring a valid topological execution order.

(2)~\textbf{Role Ordering} ($\mathcal{G}_2$): $\mathcal{G}_2 = \{G \mid \forall (v_i, v_j) \in E : \rho(v_i) \preceq \rho(v_j)\}$.

(3)~\textbf{Type Compatibility} ($\mathcal{G}_3$): $\mathcal{G}_3 = \{G \mid \forall (v_i, v_j) \in E : \tau_{\text{out}}(v_i) \subseteq \tau_{\text{in}}(v_j)\}$, where $\tau_{\text{out}}$ and $\tau_{\text{in}}$ denote output and input concept types.

(4)~\textbf{Data Availability} ($\mathcal{G}_4$): $\mathcal{G}_4 = \{G \mid G \leftrightarrow G', \forall e \in E' : e \text{ is executable}\}$.

(5)~\textbf{Connectivity} ($\mathcal{G}_5$): $\mathcal{G}_5 = \{G \mid \forall v \in V : \exists v_0 \in V_{\text{ext}}, v_m \in V_{\text{meas}} : \text{path}(v_0, v) \land \text{path}(v, v_m)\}$, where $V_{\text{ext}} = \{v : \rho(v) \in \{\textsc{Extent}, \textsc{TExtent}\}\}$ and $V_{\text{meas}} = \{v : \rho(v) = \textsc{Measure}\}$.

A GeoFlow Graph is \textit{well-formed} if it satisfies all five constraints above, i.e., $G \in \bigcap_{i=1}^{5} \mathcal{G}_i$. We formalize DAG assembly as a constraint satisfaction problem where multiple valid configurations may exist.

\subsection{Retrieval-augmented Orchestration}
\label{sec:dag-gen}

While the agent can assemble GeoFlow Graphs from primitive operators subject to the constraints defined above, we observe that geo-analytical questions exhibit recurring structural patterns. To accelerate generation and improve accuracy, we adopt a compositional approach that leverages a library of pre-validated graph templates, each of which inherently satisfies the well-formedness constraints.

\paragraph{Template Library.}
We define a set of macro-templates $\mathcal{T} = \{g_1, \ldots, g_K\}$, where each template $g_k = (V_k, E_k, \textit{in}_k, \textit{out}_k)$ specifies a subgraph $g_k \subseteq G$ with designated input and output ports. As illustrated in Figure~\ref{fig:framework} (B), the Concept Transformation Drafting stage retrieves relevant templates from the library to define transformations such as $\concept{Object} \rightarrow \concept{Field}$ and $\concept{Event} \rightarrow \concept{Proportion}$, which are then composed into an ordered GeoFlow Graph (see Appendix~\ref{app:templates} for the complete template library).

Given a question $q$, the LLM additionally generates the GeoFlow Graph guided by retrieved examples $\mathcal{E}_q$ from a question-graph store.

\subsection{Learning with Geographic Constraints}

While Spatial-Agent can operate with off-the-shelf LLMs via prompting, we optionally employ a two-stage fine-tuning strategy to further improve performance. In Stage 1, we apply Supervised Fine-Tuning (SFT) on question-concept pairs $\{(q_i, V_i)\}_{i=1}^{N}$, minimizing the negative log-likelihood $\mathcal{L}_{\text{SFT}} = -\sum_{i} \log p_\theta(V_i \mid q_i)$, where $p_\theta$ denotes the LLM parameterized by $\theta$, to train the LLM to extract spatial concepts with their types and functional roles. In Stage 2, we use Direct Preference Optimization (DPO) to train the LLM to generate well-formed GeoFlow Graphs. For each question $q$, we construct preference pairs $(G^+, G^-)$ where $G^+ \in \bigcap_{i=1}^{5} \mathcal{G}_i$ while $G^- \notin \bigcap_{i=1}^{5} \mathcal{G}_i$, and optimize:
\begin{equation}
\label{eq:dpo}
\min_{\theta} \; \mathcal{L}_{\text{DPO}}, \quad \text{s.t.} \; G^+ \in \bigcap_{i=1}^{5} \mathcal{G}_i, \; G^- \notin \bigcap_{i=1}^{5} \mathcal{G}_i
\end{equation}
where $\mathcal{L}_{\text{DPO}} = -\mathbb{E}_{(q, G^+, G^-)}\left[\log \sigma\left(\beta \cdot r_\theta\right)\right]$, $r_\theta = \log \frac{p_\theta(G^+ \mid q)}{p_{\text{ref}}(G^+ \mid q)} - \log \frac{p_\theta(G^- \mid q)}{p_{\text{ref}}(G^- \mid q)}$ is the reward margin, $\sigma$ is the sigmoid function, $\beta$ is the temperature, and $p_{\text{ref}}$ is the reference model. This optional fine-tuning enables the model to better internalize geographic reasoning patterns (see Appendix~\ref{app:fine-tuning} for details).

\subsection{Execution and Response Generation}

Given a well-formed GeoFlow Graph $G$, we first transform it to the factorized graph $G'$ and then execute transformation steps in topological order. All intermediate states are recorded and provided to the agent for response generation, ensuring the output is grounded in computational results rather than hallucinated from learned priors (see Appendix~\ref{app:execution} for the detailed algorithm).

\section{Experiments}

\subsection{Experimental Setup}

\paragraph{Datasets.}
We evaluate SpatialAgent on two comprehensive geospatial reasoning benchmarks that collectively cover diverse task types, geographic regions, and reasoning capabilities.
\textbf{MapEval-API} \cite{dihan2024mapeval} is the API-based evaluation from the MapEval benchmark, which requires agents to invoke map tools for geospatial reasoning. The dataset comprises four task categories: \textit{Place Info} (retrieving place attributes), \textit{Nearby} (finding nearby points of interest), \textit{Routing} (computing directions and distances), and \textit{Trip} (multi-stop travel planning). The benchmark spans 180 cities across 54 countries, providing diverse geographic coverage for evaluating tool-augmented spatial reasoning.
\textbf{MapQA} \cite{li2025mapqa} is an open-domain geospatial QA dataset with 3,154 question-answer pairs spanning nine question types, constructed from OpenStreetMap data covering Southern California and Illinois.

\paragraph{Baselines.}
We compare SpatialAgent against the following methods: (1) \textbf{Direct LLM}, which directly prompts the language model without agent scaffolding; (2) \textbf{ReAct} \cite{yao2022react}, an agent that interleaves reasoning and acting through thought-action-observation loops; (3) \textbf{Reflexion} \cite{shinn2023reflexion}, an agent that learns from execution failures through self-reflection and memory; and (4) \textbf{Plan-and-Solve} \cite{wang2023plan}, a prompting strategy that first generates an overall plan before execution.

\subsection{Main Results}

\paragraph{Results on MapEval-API.}
Table~\ref{tab:mapeval} presents the results on the MapEval-API benchmark across four task categories.
SpatialAgent consistently outperforms all baselines across different backbone LLMs.
For closed-source models, SpatialAgent with GPT-4o-mini achieves an overall accuracy of 45.15\%, representing a 96.30\% relative improvement over the MapEval API baseline (23.00\%).
The improvement is particularly pronounced on \textit{Place Info} (+149.91\%) and \textit{Nearby} (+133.26\%) tasks, where structured tool invocation and spatial reasoning are critical.
When equipped with GPT-5, SpatialAgent reaches the best overall accuracy of 71.88\%, with strong performance across all categories, especially on \textit{Routing} (75.76\%) and \textit{Trip} (77.61\%) tasks that require multi-step planning.

For open-source LLMs, SpatialAgent demonstrates competitive performance.
Qwen2.5-72B-Instruct achieves the best overall accuracy of 53.41\% among open-source models, with the highest \textit{Trip} accuracy (61.19\%), while Qwen2.5-32B-Instruct achieves the best \textit{Routing} accuracy (50.00\%).
With LLaMA-70B, our method achieves 47.77\% overall accuracy (+26.82\% over baseline), with substantial improvements on \textit{Place Info} (+35.29\%) and \textit{Nearby} (+62.96\%).
We observe that \textit{Routing} tasks remain challenging for LLaMA-70B (-17.85\%), suggesting that complex navigation queries benefit more from models with stronger reasoning capabilities.

\begin{table*}[t]
\centering
\small
\resizebox{\textwidth}{!}{%
\begin{tabular}{lccccc}
\toprule
\textbf{Method} & \textbf{Overall} & \textbf{Place Info} & \textbf{Nearby} & \textbf{Routing} & \textbf{Trip} \\
\midrule
\multicolumn{6}{c}{\textit{Closed-Source LLMs}} \\
\midrule
Direct LLM (GPT-4o-mini) & 32.23 & 45.31 & 32.53 & 22.73 & 28.36 \\
ReAct (GPT-4o-mini) & 32.98 & 60.94 & 22.89 & 22.73 & 25.37 \\
Reflexion (GPT-4o-mini) & 38.29 & 65.63 & 22.89 & 30.30 & 34.33 \\
MapEval API (GPT-3.5-Turbo) & 27.33 & 39.06 & 22.89 & 33.33 & 19.40 \\
MapEval API (GPT-4o-mini) & 23.00 & 28.13 & 14.46 & 13.64 & 43.28 \\
Spatial-Agent (GPT-3.5-Turbo) & 34.61 \gain{26.64\%} & 57.81 \gain{48.03\%} & 26.51 \gain{15.81\%} & \second{36.36} \gain{9.09\%} & 25.37 \gain{30.77\%} \\
Spatial-Agent (GPT-4o-mini) & \second{45.15} \gain{96.30\%} & \second{70.31} \gain{149.91\%} & \second{33.73} \gain{133.26\%} & 30.30 \gain{122.14\%} & \second{46.27} \gain{6.91\%} \\
Spatial-Agent (GPT-5) & \first{71.88} & \first{85.94} & \first{53.01} & \first{75.76} & \first{77.61} \\
\midrule
\multicolumn{6}{c}{\textit{Open-Source LLMs}} \\
\midrule
MapEval API (LlaMA-70B) & 37.67 & 53.13 & 32.53 & 42.42 & 31.34 \\
Spatial-Agent (LlaMA-70B) & 47.77 \gain{26.82\%} & \first{71.88} \gain{35.29\%} & \first{53.01} \gain{62.96\%} & 34.85 \drop{17.85\%} & 31.34 \gain{0.00\%} \\
Spatial-Agent (Qwen2.5-72B-Instruct) & \first{53.41} & \second{68.75} & \second{39.76} & \second{43.94} & \first{61.19} \\
Spatial-Agent (Qwen2.5-32B-Instruct) & \second{52.35} & \first{71.88} & \second{39.76} & \first{50.00} & \second{47.76} \\
Spatial-Agent (Gemma-2-9B) & 22.70 & 31.25 & 14.46 & 19.70 & 25.37 \\
\bottomrule
\end{tabular}
}
\caption{Results on MapEval-API. Accuracy (\%) across four task categories. \first{Bold}/\second{underline}: best/second-best. \textcolor{blue}{$\uparrow$}/\textcolor{red}{$\downarrow$}: improvement/decrease over baseline.}
\label{tab:mapeval}
\end{table*}

\paragraph{Results on MapQA.}
Table~\ref{tab:mapqa} presents the results on the MapQA benchmark across six question types.
For closed-source models, SpatialAgent (GPT-4o-mini) achieves the best overall accuracy of 61.45\%, substantially outperforming Direct LLM (13.55\%), ReAct (43.79\%), and Reflexion (53.79\%).
The improvement is particularly notable on \textit{Amenities-Around} and \textit{Amenities-Around-Specific} tasks, demonstrating the effectiveness of our spatial reasoning framework for complex location-based queries.

For open-source models, SpatialAgent with LLaMA-70B achieves the highest overall accuracy of 62.45\%, with the best performance on \textit{Amenities} (84.00\%).
Qwen2.5-72B-Instruct achieves comparable overall accuracy (61.45\%) and the highest \textit{Amenities-Around-Specific} accuracy (78.00\%).
Notably, the open-source SpatialAgent variants achieve competitive or superior performance compared to the closed-source GPT-4o-mini configuration, demonstrating the generalizability of our framework across different model families.

\begin{table*}[t]
\centering
\small
\resizebox{\textwidth}{!}{%
\begin{tabular}{clccccccc}
\toprule
\textbf{Type} & \textbf{Method}
& \textbf{Overall}
& \textbf{Adj}
& \textbf{Amen}
& \textbf{Amen-A}
& \textbf{Amen-AS}
& \textbf{Cmp-Cl}
& \textbf{Dist} \\
\midrule
\multirow{4}{*}{\textit{Closed-Source}} & Direct LLM (GPT-4o-mini)      & 13.55 & 6.00 & 36.00 & 4.00 & 0.00 & \second{35.29} & 0.00 \\
& ReAct (GPT-4o-mini)           & 43.79 & \first{56.00} & 28.00 & 16.00 & \second{64.00} & \first{64.71} & \second{34.00} \\
& Reflexion (GPT-4o-mini)       & \second{53.79} & 50.00 & \second{80.00} & \second{30.00} & \second{64.00} & \first{64.71} & \second{34.00} \\
& Spatial-Agent (GPT-4o-mini)   & \first{61.45} & \second{52.00} & \first{82.00} & \first{56.00} & \first{74.00} & \first{64.71} & \first{40.00} \\
\midrule
\multirow{2}{*}{\textit{Open-Source}} & Spatial-Agent (Qwen2.5-72B-Instruct)          & 61.45 & 48.00 & 76.00 & 64.00 & 78.00 & 64.71 & 38.00 \\
& Spatial-Agent (LlaMA-70B)         & 62.45 & 54.00 & 84.00 & 58.00 & 76.00 & 64.71 & 38.00 \\
\bottomrule
\end{tabular}
}
\caption{Results on the MapQA benchmark. 
Adj: Adjacent; Amen: Amenities; Amen-A: Amenities-Around; Amen-AS: Amenities-Around-Specific; 
Cmp-Cl: Compare-Closer; Dist: Distance.}
\label{tab:mapqa}
\end{table*}

\subsection{Analysis}
\begin{figure}[t]
    \centering
    \includegraphics[width=0.7\columnwidth]{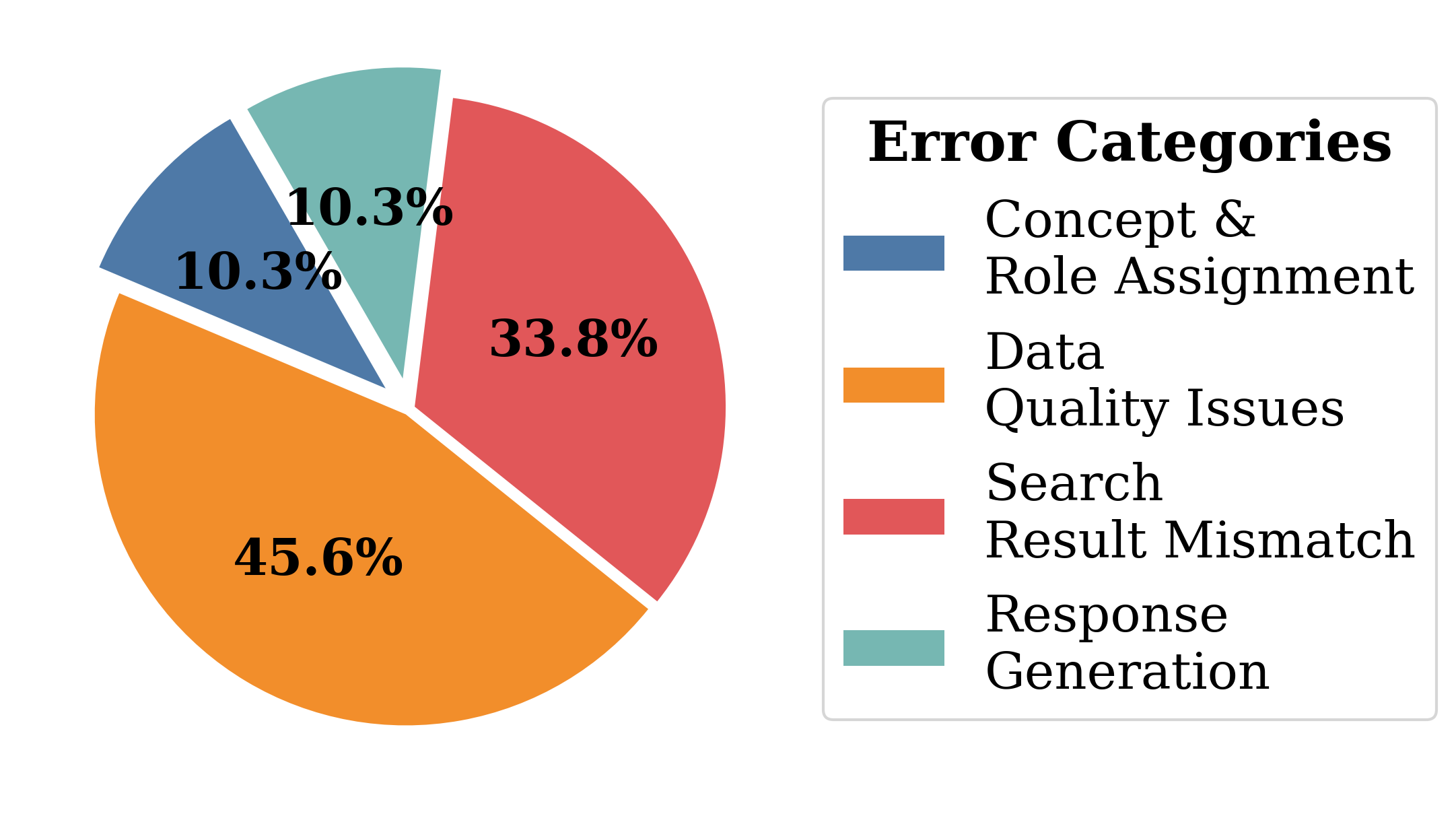}
    \caption{Distribution of error categories in SpatialAgent. Data Quality Issues (45.6\%) and Search Result Mismatch (33.8\%) account for the majority of errors, both occurring during the execution stage.}
    \label{fig:error-analysis}
\end{figure}

\paragraph{Error Analysis.}
To understand the limitations of SpatialAgent, we manually analyzed 68 incorrect predictions on MapEval-API, categorizing errors according to our four-stage pipeline (Figure~\ref{fig:error-analysis}).

We categorize errors into four types (Figure~\ref{fig:error-analysis}): Data Quality Issues (45.6\%) and Search Result Mismatch (33.8\%) occur during execution when external APIs return incomplete data or mismatched results; Concept \& Role Assignment (10.3\%) involves misidentifying concepts or functional roles; Response Generation (10.3\%) occurs when correct execution leads to incorrect answer selection. Notably, no errors originated from GeoFlow Graph construction itself, validating our template-based approach. This confirms that the primary bottleneck lies in external API interactions rather than reasoning components.

\begin{figure}[t]
\centering
\includegraphics[width=0.9\linewidth]{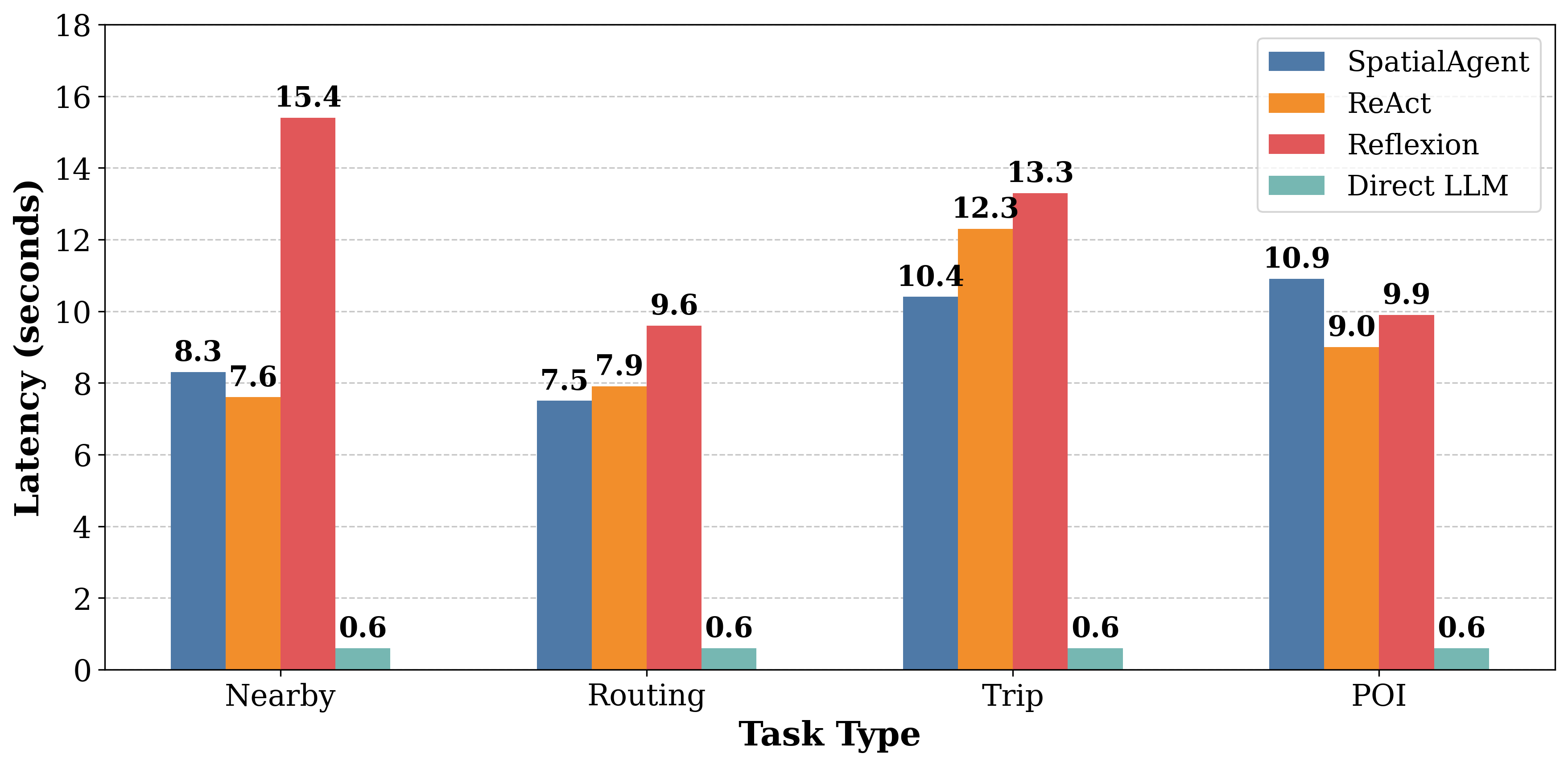}
\caption{Average latency per query (seconds) across task types. All methods use GPT-4o-mini.}
\label{fig:latency}
\end{figure}

\paragraph{Latency Analysis.}
Figure~\ref{fig:latency} compares response latency across methods using GPT-4o-mini. Direct LLM achieves the lowest latency (0.6s) but poor accuracy due to lack of grounding. Among agentic methods, SpatialAgent shows competitive latency: fastest on Routing (7.5s), comparable to ReAct on Nearby (8.3s vs 7.6s) and Trip (10.4s vs 12.3s), slightly slower on POI (10.9s vs 9.0s). Reflexion consistently exhibits the highest latency due to its iterative self-reflection mechanism.

\paragraph{Cost Analysis.}
We compare token consumption across methods: SpatialAgent uses 9,185 input and 1,451 output tokens per query; Reflexion consumes the most (10,964 input, 1,271 output); ReAct uses fewer (7,354 input, 735 output); Direct LLM is minimal (1,500 input, 50 output). With GPT-4o-mini pricing (\$0.15/1M input, \$0.60/1M output), all methods cost below \$0.003 per query, with SpatialAgent at \$0.0022 offering the best accuracy-cost trade-off.

\subsection{Ablation Study}

\paragraph{Effect of Fine-tuning.}
We evaluate our two-stage fine-tuning on Qwen-14B (Table~\ref{tab:ablation}, bottom). The base model achieves 49.59\% accuracy. SFT alone improves to 56.84\% (+14.6\%) by learning concept extraction and role assignment; DPO alone yields 55.13\% (+11.2\%) through preference learning for graph construction. However, both single-stage strategies cause drops on \textit{Route} tasks, suggesting overfitting on certain patterns. The combined SFT+DPO achieves the best accuracy of 60.58\% (+22.2\%) with consistent improvements across all categories, especially \textit{Trip} (+50.0\%), indicating that multi-step planning benefits from both accurate concept extraction and valid graph composition.

\begin{table}[t]
    \centering
    \resizebox{\columnwidth}{!}{%
    \begin{tabular}{lccccc}
    \toprule
    \textbf{Method} & \textbf{All} & \textbf{POI} & \textbf{Near} & \textbf{Route} & \textbf{Trip} \\
    \midrule
    Plan-and-Solve (4o-mini) & \second{41.17} & \second{60.94} & \second{30.12} & \first{31.82} & 41.79 \\
    Spatial-Agent (4o-mini) & \first{45.15} & \first{70.31} & \first{33.73} & \second{30.30} & \first{46.27} \\
    \quad w/o Template & 39.32 \drop{12.9\%} & 57.81 \drop{17.8\%} & 28.92 \drop{14.3\%} & 27.27 \drop{10.0\%} & \second{43.28} \drop{6.5\%} \\
    \midrule
    Spatial-Agent (Qwen-14B) & 49.59 & 64.06 & \second{46.99} & \second{48.48} & 38.81 \\
    \quad w/ SFT only & \second{56.84} \gain{14.6\%} & 79.69 \gain{24.4\%} & \second{46.99} \gain{0.0\%} & 45.45 \drop{6.3\%} & \second{55.22} \gain{42.3\%} \\
    \quad w/ DPO only & 55.13 \gain{11.2\%} & \second{81.25} \gain{26.8\%} & 44.58 \drop{5.1\%} & 43.94 \drop{9.4\%} & 50.75 \gain{30.8\%} \\
    \quad w/ SFT + DPO & \first{60.58} \gain{22.2\%} & \first{84.38} \gain{31.7\%} & \first{48.19} \gain{2.6\%} & \first{51.52} \gain{6.3\%} & \first{58.21} \gain{50.0\%} \\
    \bottomrule
    \end{tabular}
    }
    \caption{Ablation study results.}
    \label{tab:ablation}
\end{table}

\paragraph{Effect of Template Composition.}
We compare Spatial-Agent with and without templates using GPT-4o-mini (Table~\ref{tab:ablation}, top). Removing templates drops accuracy from 45.15\% to 39.32\% (-12.9\%), with consistent degradation across categories: \textit{POI} (-17.8\%), \textit{Near} (-14.3\%), \textit{Route} (-10.0\%), \textit{Trip} (-6.5\%). This validates that templates encoding recurring geo-analytical patterns improve validity.

\section{Conclusion}
We present Spatial-Agent, a geospatial AI agent that formalizes geo-analytical question answering as a concept transformation problem grounded in GIScience theory. The GeoFlow Graph representation encodes spatial concepts, functional roles, and well-formedness constraints, enabling structured reasoning over geographic workflows. Our template-based compositional generation leverages recurring geo-analytical patterns to improve structural validity, while SFT+DPO fine-tuning enables models to internalize geographic constraints.
Experiments demonstrate that Spatial-Agent significantly outperforms existing agent baselines. Error analysis reveals that the primary bottleneck lies in external API interactions rather than reasoning components, validating the effectiveness of our structured spatial reasoning approach.

\section*{Limitations}
Several limitations remain. First, the framework's accuracy is bounded by the reliability of external geospatial APIs, with most errors occurring during execution due to data quality issues. Second, the template library may not cover all question types, requiring from-scratch generation for novel patterns. Third, the fine-tuning approach requires annotated data that demands significant effort to scale to new domains. Finally, our evaluation focuses on English-language urban environments; performance on specialized geographic domains remains unexplored.

\bibliography{custom}

@article{yu2025spatial,
  title={Spatial-rag: Spatial retrieval augmented generation for real-world spatial reasoning questions},
  author={Yu, Dazhou and Bao, Riyang and Mai, Gengchen and Zhao, Liang},
  journal={arXiv preprint arXiv:2502.18470},
  year={2025}
}

@article{goodchild1992geographical,
  title={Geographical information science},
  author={Goodchild, Michael F},
  journal={International journal of geographical information systems},
  volume={6},
  number={1},
  pages={31--45},
  year={1992},
  publisher={Taylor \& Francis}
}

@article{kuhn2012core,
  title={Core concepts of spatial information for transdisciplinary research},
  author={Kuhn, Werner},
  journal={International Journal of Geographical Information Science},
  volume={26},
  number={12},
  pages={2267--2276},
  year={2012},
  publisher={Taylor \& Francis}
}

@article{mai2025towards,
  title={Towards the next generation of Geospatial Artificial Intelligence},
  author={Mai, Gengchen and Xie, Yiqun and Jia, Xiaowei and Lao, Ni and Rao, Jinmeng and Zhu, Qing and Liu, Zeping and Chiang, Yao-Yi and Jiao, Junfeng},
  journal={International Journal of Applied Earth Observation and Geoinformation},
  volume={136},
  pages={104368},
  year={2025},
  publisher={Elsevier}
}

@article{li2021neural,
  title={Neural factoid geospatial question answering},
  author={Li, Haonan and Hamzei, Ehsan and Majic, Ivan and Hua, Hua and Renz, Jochen and Tomko, Martin and Vasardani, Maria and Winter, Stephan and Baldwin, Timothy},
  journal={Journal of Spatial Information Science},
  number={23},
  pages={65--90},
  year={2021}
}

@article{scheider2021geo,
  title={Geo-analytical question-answering with GIS},
  author={Scheider, Simon and Nyamsuren, Enkhbold and Kruiger, Han and Xu, Haiqi},
  journal={International Journal of Digital Earth},
  volume={14},
  number={1},
  pages={1--14},
  year={2021},
  publisher={Taylor \& Francis}
}

@article{tobler1970computer,
  title={A computer movie simulating urban growth in the Detroit region},
  author={Tobler, Waldo R},
  journal={Economic geography},
  volume={46},
  number={sup1},
  pages={234--240},
  year={1970},
  publisher={Taylor \& Francis}
}

@article{janowicz2012geospatial,
  title={Geospatial semantics and linked spatiotemporal data--Past, present, and future},
  author={Janowicz, Krzysztof and Scheider, Simon and Pehle, Todd and Hart, Glen},
  journal={Semantic Web},
  volume={3},
  number={4},
  pages={321--332},
  year={2012},
  publisher={SAGE Publications Sage UK: London, England}
}

@article{janowicz2012observation,
  title={Observation-driven geo-ontology engineering},
  author={Janowicz, Krzysztof},
  journal={Transactions in GIS},
  volume={16},
  number={3},
  pages={351--374},
  year={2012},
  publisher={Wiley Online Library}
}

@article{janowicz2022know,
  title={Know, Know Where, KnowWhereGraph: A densely connected, cross-domain knowledge graph and geo-enrichment service stack for applications in environmental intelligence},
  author={Janowicz, Krzysztof and Hitzler, Pascal and Li, Wenwen and Rehberger, Dean and Schildhauer, Mark and Zhu, Rui and Shimizu, Cogan and Fisher, Colby and Cai, Ling and Mai, Gengchen and others},
  journal={AI Magazine},
  volume={43},
  number={1},
  pages={30--39},
  year={2022}
}

@article{mai2021geographic,
  title={Geographic question answering: challenges, uniqueness, classification, and future directions},
  author={Mai, Gengchen and Janowicz, Krzysztof and Zhu, Rui and Cai, Ling and Lao, Ni},
  journal={AGILE: GIScience series},
  volume={2},
  pages={8},
  year={2021},
  publisher={Copernicus Publications G{\"o}ttingen, Germany}
}

@inproceedings{li2023location,
  title={Location Aware Modular Biencoder for Tourism Question Answering},
  author={Li, Haonan and Tomko, Martin and Baldwin, Timothy},
  booktitle={Findings of the Association for Computational Linguistics: IJCNLP-AACL 2023 (Findings)},
  pages={95--109},
  year={2023}
}

@inproceedings{contractor2021joint,
  title={Joint spatio-textual reasoning for answering tourism questions},
  author={Contractor, Danish and Goel, Shashank and Mausam and Singla, Parag},
  booktitle={Proceedings of the Web Conference 2021},
  pages={1978--1989},
  year={2021}
}

@inproceedings{contractor2021answering,
  title={Answering POI-recommendation questions using tourism reviews},
  author={Contractor, Danish and Shah, Krunal and Partap, Aditi and Singla, Parag and Mausam, Mausam},
  booktitle={Proceedings of the 30th ACM International Conference on Information \& Knowledge Management},
  pages={281--291},
  year={2021}
}

@article{mai2020se,
  title={SE-KGE: A location-aware knowledge graph embedding model for geographic question answering and spatial semantic lifting},
  author={Mai, Gengchen and Janowicz, Krzysztof and Cai, Ling and Zhu, Rui and Regalia, Blake and Yan, Bo and Shi, Meilin and Lao, Ni},
  journal={Transactions in GIS},
  volume={24},
  number={3},
  pages={623--655},
  year={2020},
  publisher={Wiley Online Library}
}

@article{goodchild2007towards,
  title={Towards a general theory of geographic representation in GIS},
  author={Goodchild, Michael F and Yuan, May and Cova, Thomas J},
  journal={International journal of geographical information science},
  volume={21},
  number={3},
  pages={239--260},
  year={2007},
  publisher={Taylor \& Francis}
}

@article{miller2015data,
  title={Data-driven geography},
  author={Miller, Harvey J and Goodchild, Michael F},
  journal={GeoJournal},
  volume={80},
  number={4},
  pages={449--461},
  year={2015},
  publisher={Springer}
}

@article{scheider2020ontology,
  title={Ontology of core concept data types for answering geo-analytical questions},
  author={Scheider, Simon and Meerlo, Rogier and Kasalica, Vedran and Lamprecht, Anna-Lena},
  journal={Journal of Spatial Information Science},
  number={20},
  pages={167--201},
  year={2020}
}

@article{xu2023grammar,
  title={A grammar for interpreting geo-analytical questions as concept transformations},
  author={Xu, Haiqi and Nyamsuren, Enkhbold and Scheider, Simon and Top, Eric},
  journal={International Journal of Geographical Information Science},
  volume={37},
  number={2},
  pages={276--306},
  year={2023},
  publisher={Taylor \& Francis}
}

@inproceedings{punjani2018template,
  title={Template-based question answering over linked geospatial data},
  author={Punjani, Dharmen and Singh, Kuldeep and Both, Andreas and Koubarakis, Manolis and Angelidis, Iosif and Bereta, Konstantina and Beris, Themis and Bilidas, Dimitris and Ioannidis, Theofilos and Karalis, Nikolaos and others},
  booktitle={Proceedings of the 12th workshop on geographic information retrieval},
  pages={1--10},
  year={2018}
}

@article{kefalidis2024question,
  title={The question answering system GeoQA2 and a new benchmark for its evaluation},
  author={Kefalidis, Sergios-Anestis and Punjani, Dharmen and Tsalapati, Eleni and Plas, Konstantinos and Pollali, Maria-Aggeliki and Maret, Pierre and Koubarakis, Manolis},
  journal={International Journal of Applied Earth Observation and Geoinformation},
  volume={134},
  pages={104203},
  year={2024},
  publisher={Elsevier}
}

@inproceedings{kefalidis2023benchmarking,
  title={Benchmarking geospatial question answering engines using the dataset GeoQuestions1089},
  author={Kefalidis, Sergios-Anestis and Punjani, Dharmen and Tsalapati, Eleni and Plas, Konstantinos and Pollali, Mariangela and Mitsios, Michail and Tsokanaridou, Myrto and Koubarakis, Manolis and Maret, Pierre},
  booktitle={International semantic web conference},
  pages={266--284},
  year={2023},
  organization={Springer}
}

@article{li2025mapqa,
  title={MapQA: Open-domain Geospatial Question Answering on Map Data},
  author={Li, Zekun and Grossman, Malcolm and Kulkarni, Mihir and Chen, Muhao and Chiang, Yao-Yi and others},
  journal={arXiv preprint arXiv:2503.07871},
  year={2025}
}

@inproceedings{yao2022react,
  title={React: Synergizing reasoning and acting in language models},
  author={Yao, Shunyu and Zhao, Jeffrey and Yu, Dian and Du, Nan and Shafran, Izhak and Narasimhan, Karthik R and Cao, Yuan},
  booktitle={The eleventh international conference on learning representations},
  year={2022}
}

@article{schick2023toolformer,
  title={Toolformer: Language models can teach themselves to use tools},
  author={Schick, Timo and Dwivedi-Yu, Jane and Dess{\`\i}, Roberto and Raileanu, Roberta and Lomeli, Maria and Hambro, Eric and Zettlemoyer, Luke and Cancedda, Nicola and Scialom, Thomas},
  journal={Advances in Neural Information Processing Systems},
  volume={36},
  pages={68539--68551},
  year={2023}
}

@inproceedings{wang2024executable,
  title={Executable code actions elicit better llm agents},
  author={Wang, Xingyao and Chen, Yangyi and Yuan, Lifan and Zhang, Yizhe and Li, Yunzhu and Peng, Hao and Ji, Heng},
  booktitle={Forty-first International Conference on Machine Learning},
  year={2024}
}

@article{li2023autonomous,
  title={Autonomous GIS: the next-generation AI-powered GIS},
  author={Li, Zhenlong and Ning, Huan},
  journal={International Journal of Digital Earth},
  volume={16},
  number={2},
  pages={4668--4686},
  year={2023},
  publisher={Taylor \& Francis}
}

@article{zhang2024geogpt,
  title={GeoGPT: An assistant for understanding and processing geospatial tasks},
  author={Zhang, Yifan and Wei, Cheng and He, Zhengting and Yu, Wenhao},
  journal={International Journal of Applied Earth Observation and Geoinformation},
  volume={131},
  pages={103976},
  year={2024},
  publisher={Elsevier}
}

@article{chen2024llm,
  title={An llm agent for automatic geospatial data analysis},
  author={Chen, Yuxing and Wang, Weijie and Lobry, Sylvain and Kurtz, Camille},
  journal={arXiv preprint arXiv:2410.18792},
  year={2024}
}

@article{zhang2025geospatial,
  title={Geospatial large language model trained with a simulated environment for generating tool-use chains autonomously},
  author={Zhang, Yifan and Li, Jingxuan and Wang, Zhiyun and He, Zhengting and Guan, Qingfeng and Lin, Jianfeng and Yu, Wenhao},
  journal={International Journal of Applied Earth Observation and Geoinformation},
  volume={136},
  pages={104312},
  year={2025},
  publisher={Elsevier}
}

@article{kruiger2021loose,
  title={Loose programming of GIS workflows with geo-analytical concepts},
  author={Kruiger, Johannes F and Kasalica, Vedran and Meerlo, Rogier and Lamprecht, Anna-Lena and Nyamsuren, Enkhbold and Scheider, Simon},
  journal={Transactions in GIS},
  volume={25},
  number={1},
  pages={424--449},
  year={2021},
  publisher={Wiley Online Library}
}

@inproceedings{devlin2017robustfill,
  title={Robustfill: Neural program learning under noisy i/o},
  author={Devlin, Jacob and Uesato, Jonathan and Bhupatiraju, Surya and Singh, Rishabh and Mohamed, Abdel-rahman and Kohli, Pushmeet},
  booktitle={International conference on machine learning},
  pages={990--998},
  year={2017},
  organization={PMLR}
}

@article{zhong2023hierarchical,
  title={Hierarchical neural program synthesis},
  author={Zhong, Linghan and Lindeborg, Ryan and Zhang, Jesse and Lim, Joseph J and Sun, Shao-Hua},
  journal={arXiv preprint arXiv:2303.06018},
  year={2023}
}

@article{dihan2024mapeval,
  title={Mapeval: A map-based evaluation of geo-spatial reasoning in foundation models},
  author={Dihan, Mahir Labib and Hassan, Md Tanvir and Parvez, Md Tanvir and Hasan, Md Hasebul and Alam, Md Almash and Cheema, Muhammad Aamir and Ali, Mohammed Eunus and Parvez, Md Rizwan},
  journal={arXiv preprint arXiv:2501.00316},
  year={2024}
}

@article{shinn2023reflexion,
  title={Reflexion: Language agents with verbal reinforcement learning},
  author={Shinn, Noah and Cassano, Federico and Gopinath, Ashwin and Narasimhan, Karthik and Yao, Shunyu},
  journal={Advances in Neural Information Processing Systems},
  volume={36},
  pages={8634--8652},
  year={2023}
}

@article{wang2023plan,
  title={Plan-and-solve prompting: Improving zero-shot chain-of-thought reasoning by large language models},
  author={Wang, Lei and Xu, Wanyu and Lan, Yihuai and Hu, Zhiqiang and Lan, Yunshi and Lee, Roy Ka-Wei and Lim, Ee-Peng},
  journal={arXiv preprint arXiv:2305.04091},
  year={2023}
}

@article{yan2023inherent,
  title={Inherent limitations of LLMs regarding spatial information},
  author={Yan, He and Hu, Xinyao and Wan, Xiangpeng and Huang, Chengyu and Zou, Kai and Xu, Shiqi},
  journal={arXiv preprint arXiv:2312.03042},
  year={2023}
}

@inproceedings{wang2025geohallucination,
  title={Mitigating Geospatial Knowledge Hallucination in Large Language Models: Benchmarking and Dynamic Factuality Aligning},
  author={Wang, Shengyuan and Feng, Jie and Liu, Tianhui and Pei, Dan and Li, Yong},
  booktitle={Findings of the Association for Computational Linguistics: EMNLP 2025},
  pages={870--888},
  year={2025}
}

@article{zhang2025geoanalystbench,
  title={GeoAnalystBench: A GeoAI benchmark for assessing large language models for spatial analysis workflow and code generation},
  author={Zhang, Qianheng and Gao, Song and Wei, Chen and Zhao, Yibo and Nie, Ying and Chen, Ziru and Chen, Shijie and Su, Yu and Sun, Huan},
  journal={Transactions in GIS},
  volume={29},
  number={7},
  pages={e70135},
  year={2025},
  publisher={Wiley Online Library}
}

@article{ji2025foundation,
  title={Foundation models for geospatial reasoning: assessing the capabilities of large language models in understanding geometries and topological spatial relations},
  author={Ji, Yuhan and Gao, Song and Nie, Ying and Maji{\'c}, Ivan and Janowicz, Krzysztof},
  journal={International Journal of Geographical Information Science},
  pages={1--38},
  year={2025},
  publisher={Taylor \& Francis}
}

\clearpage
\appendix

\section{Core Spatial Concepts}
\label{app:core-concepts}

Following Kuhn's theory of core concepts in spatial information~\cite{kuhn2012core},
we define a fixed set of primitive spatial concepts that constitute the semantic
foundation of geo-analytical reasoning.

\subsection{Definition}

\begin{definition}[Core Spatial Concept Space]
The spatial concept space is defined as:
\begin{equation}
\mathcal{C} = \left\{\begin{aligned}
  &\concept{Location}, \concept{Object}, \concept{Field}, \concept{Event},\\
  &\concept{Network}, \concept{Amount}, \concept{Proportion}
\end{aligned}\right\}.
\end{equation}
Each concept $c \in \mathcal{C}$ represents a fundamental category of geographic
phenomena and serves as an atomic unit in concept transformation reasoning.
\end{definition}

\subsection{Concept Semantics}

Each core concept is associated with a distinct ontological interpretation:

\begin{itemize}
  \item \textsc{Location}: A spatial reference or place identifier, typically used
  to anchor other concepts (e.g., cities, regions, coordinates).
  \item \textsc{Object}: Discrete spatial entities with identifiable boundaries
  (e.g., buildings, schools, fire stations).
  \item \textsc{Field}: Continuous spatial distributions over space
  (e.g., elevation, temperature, distance fields).
  \item \textsc{Event}: Temporally bounded spatial occurrences
  (e.g., traffic accidents, crime incidents).
  \item \textsc{Network}: Graph-structured spatial systems supporting connectivity
  and routing (e.g., road networks, river networks).
  \item \textsc{Amount}: Quantitative aggregations derived from objects, fields,
  or events (e.g., population count, total area).
  \item \textsc{Proportion}: Normalized quantities expressing ratios or densities
  (e.g., population density, percentage coverage).
\end{itemize}

\subsection{Examples}

Table~\ref{tab:core-concept-examples} illustrates how natural-language phrases are
mapped to core spatial concepts.

\begin{table}[htbp]
\centering
\small
\begin{tabular}{ll}
\hline
\textbf{Natural-language phrase} & \textbf{Core concept} \\
\hline
``fire stations'' & \textsc{Object} \\
``Euclidean distance'' & \textsc{Field} \\
``traffic accidents in 2025'' & \textsc{Event} \\
``road network'' & \textsc{Network} \\
``total population'' & \textsc{Amount} \\
``population density'' & \textsc{Proportion} \\
\hline
\end{tabular}
\caption{Examples of core spatial concepts extracted from natural-language questions.}
\label{tab:core-concept-examples}
\end{table}

\section{Functional Roles in Geo-Analytical Questions}
\label{app:functional-roles}

To explicitly represent the procedural structure embedded in geo-analytical
questions, we adopt a set of functional roles inspired by the GeoAnQu framework~\cite{xu2023grammar}.

\subsection{Definition}

\begin{definition}[Functional Role Set]
We define the functional role space as:
\begin{equation}
\mathcal{R} = \left\{\begin{aligned}
  &\concept{Extent}, \concept{TExtent}, \concept{SubCond},\\
  &\concept{Cond}, \concept{Support}, \concept{Measure}
\end{aligned}\right\}.
\end{equation}
Each role $r \in \mathcal{R}$ specifies how a concept participates in a
geo-analytical workflow.
\end{definition}

\subsection{Contextual vs. Procedural Roles}

Functional roles are divided into two categories:

\paragraph{Contextual Roles}
\begin{itemize}
  \item \textsc{Extent}: Spatial scope restricting data collection
  (e.g., ``in New York City'').
  \item \textsc{TExtent}: Temporal scope restricting data collection
  (e.g., ``in 2025'').
\end{itemize}

Contextual roles constrain the domain of analysis but do not participate in
procedural ordering.

\paragraph{Procedural Roles}
\begin{itemize}
  \item \textsc{SubCond}: Preliminary constraints that restrict candidate entities
  (e.g., ``within 500 meters of rivers'').
  \item \textsc{Cond}: Conditions that further filter or transform intermediate
  results.
  \item \textsc{Support}: Spatial structures or reference objects used to compute
  derived concepts (e.g., road networks, buffers).
  \item \textsc{Measure}: The target output of the analysis.
\end{itemize}

\subsection{Procedural Precedence}

Procedural roles follow a strict execution order defined as:
\begin{equation}
\concept{SubCond} \prec \concept{Cond} \prec \concept{Support} \prec \concept{Measure}.
\end{equation}

This ordering reflects the semantics of geo-analytical workflows: constraints
are applied before supports are constructed, and measurements are computed last.

\subsection{Examples}

Table~\ref{tab:functional-role-examples} shows functional role assignments for a
representative geo-analytical question: ``How many restaurants within 500m of coffee shops near parks along subway lines in San Francisco opened last week?''

\begin{table}[htbp]
\centering
\small
\begin{tabular}{p{3.8cm}ll}
\hline
\textbf{Phrase} & \textbf{Concept} & \textbf{Role} \\
\hline
``parks'' & \textsc{Object} & \textsc{SubCond} \\
``within 500m of coffee shops'' & \textsc{Object} & \textsc{Cond} \\
``subway lines'' & \textsc{Network} & \textsc{Support} \\
``restaurants'' & \textsc{Object} & \textsc{Measure} \\
``in San Francisco'' & \textsc{Location} & \textsc{Extent} \\
``last week'' & \textsc{Event} & \textsc{TExtent} \\
\hline
\end{tabular}
\caption{Examples of functional role assignments in a geo-analytical question.}
\label{tab:functional-role-examples}
\end{table}

\section{Operator Library}
\label{app:operators}

We maintain a library of atomic operators spanning several functional categories. We detail representative operators below.

\subsection{Geocoding and Location Resolution}

\paragraph{geocode($t$, $a$, $r$)} Converts a textual address or place name $t$ into geographic coordinates $(\phi, \lambda)$. An optional anchor location $a$ provides disambiguation bias when multiple candidates exist. The region hint $r$ (ISO 3166-1 alpha-2 code) further constrains the search space. When the primary geocoding API fails, a progressive fallback strategy employs nearby search with expanding radii (10km $\to$ 50km $\to$ 100km).

\paragraph{batch\_geocode($T$, $a$)} Batch variant that processes a list of place names $T = \{t_1, \ldots, t_n\}$ with shared anchor bias, returning a list of resolved locations.

\paragraph{reverse\_geocode($\phi$, $\lambda$)} Inverse operation that converts coordinates to a human-readable address string.

\subsection{Place Search and Retrieval}

\paragraph{place\_search($c$, $\rho$, $\tau$, $\kappa$, $r_{\min}$)} Performs a spatial search centered at location $c$ within radius $\rho$ meters. Optional filters include place type $\tau$ (e.g., \texttt{restaurant}, \texttt{hospital}), keyword $\kappa$, minimum rating threshold $r_{\min}$, and current operating status.

\paragraph{place\_details($p$)} Retrieves comprehensive metadata for place $p$, including name, coordinates, rating, price level, opening hours (structured as weekly periods), phone number, and associated place types.

\paragraph{batch\_place\_details($P$)} Batch variant that enriches a list of places $P$ with detailed metadata, merging results with existing attributes.

\subsection{Routing and Navigation}

\paragraph{directions($o$, $d$, $m$, $W$)} Computes a route from origin $o$ to destination $d$ using travel mode $m \in \{\texttt{driving}, \texttt{walking}, \texttt{transit}, \texttt{bicycling}\}$. Optional waypoints $W = \{w_1, \ldots, w_k\}$ are automatically geocoded if provided as place names. Returns structured route data including legs, steps, distance, and duration. A waypoint verification mechanism checks whether specified intermediate stops are reflected in the returned route.

\paragraph{distance\_matrix($O$, $D$, $m$)} Computes a $|O| \times |D|$ matrix of travel distances and durations between origin set $O$ and destination set $D$ using travel mode $m$.

\paragraph{compare\_routes($R$, $\mu$)} Compares multiple candidate routes $R = \{r_1, \ldots, r_n\}$ and returns the index of the optimal route according to metric $\mu \in \{\texttt{distance}, \texttt{duration}\}$.

\paragraph{filter\_routes($R$, $\kappa$)} Filters routes based on instruction content, returning the index of routes containing (or avoiding) specific features indicated by keyword $\kappa$ (e.g., \texttt{stairs}, \texttt{toll}, \texttt{roundabout}).

\paragraph{extract\_distance($r$) / extract\_duration($r$)} Utility operators that extract aggregate distance (meters) or duration (seconds) from a route object $r$.

\subsection{Geometric Computation}

\paragraph{haversine($\phi_1$, $\lambda_1$, $\phi_2$, $\lambda_2$)} Computes the great-circle distance between two points on Earth's surface using the Haversine formula:
\begin{equation}
d = 2R \cdot \arctan2\left(\sqrt{a}, \sqrt{1-a}\right)
\end{equation}
where $a = \sin^2\left(\frac{\Delta\phi}{2}\right) + \cos(\phi_1)\cos(\phi_2)\sin^2\left(\frac{\Delta\lambda}{2}\right)$ and $R = 6371$ km is Earth's mean radius.

\paragraph{bearing($\phi_1$, $\lambda_1$, $\phi_2$, $\lambda_2$)} Computes the initial bearing (forward azimuth) from point 1 to point 2, returned as degrees clockwise from true north ($0^\circ$--$360^\circ$).

\paragraph{bearing\_to\_direction($\theta$)} Converts a numeric bearing $\theta$ to a cardinal/intercardinal direction string from the set $\{$N, NE, E, SE, S, SW, W, NW$\}$.

\subsection{Spatial Analysis}

\paragraph{nearest($a$, $C$, $\mu$)} Finds the candidate $c^* \in C$ that minimizes distance to anchor point $a$. The metric $\mu$ can be \texttt{haversine} (geodesic distance) or \texttt{travel\_time} (network-based).

\paragraph{within\_radius($c$, $\rho$, $C$)} Filters candidates $C$ to return only those within radius $\rho$ meters of center point $c$.

\paragraph{pairwise\_extremes($L$)} Identifies the pair of locations $(l_i, l_j) \in L \times L$ with maximum mutual distance, useful for determining spatial extent.

\paragraph{filter\_places($P$, $\theta$)} Filters a place list $P$ according to constraints $\theta$, which may include minimum rating, price level, required types, and operating status.

\subsection{Temporal Reasoning}

\paragraph{open\_at\_time($p$, $t$)} Determines whether place $p$ is open at local datetime $t$ by parsing structured opening hours. Handles edge cases including cross-midnight periods (e.g., 23:00--02:00) and 24-hour establishments.

\paragraph{timezone($\phi$, $\lambda$, $\tau$)} Retrieves timezone information for coordinates $(\phi, \lambda)$ at Unix timestamp $\tau$, returning timezone ID, name, and UTC offset.

\paragraph{calculate\_finish\_time($t_0$, $L$, $S$, $m$)} Computes the finish time of a multi-stop itinerary starting at time $t_0$, visiting locations $L$ with stay durations $S$ using travel mode $m$. Automatically queries travel times between consecutive stops.

\subsection{Trip Optimization}

\paragraph{tsp\_tw($D$, $L$, $S$, $W$, $t_0$, $T$)} Solves the Traveling Salesman Problem with Time Windows (TSP-TW) using Google OR-Tools. Given distance matrix $D$, locations $L$, service times $S$, time windows $W$, start time $t_0$, and time budget $T$, returns an optimized visit sequence that minimizes total travel time while respecting constraints.

When the complete solution violates time constraints, a greedy fallback algorithm constructs a partial feasible solution by iteratively adding the nearest unvisited location until the budget is exhausted.

\paragraph{steps\_analysis($r$, $\ell$)} Analyzes route instructions to extract navigation statistics: counts of left/right turns, roundabout exits, and optionally the instruction immediately following a specified landmark $\ell$.

\subsection{Local Context Operators}

For efficiency, we maintain a local context database that caches frequently accessed data. Six operators provide database-first retrieval with API fallback:

\begin{itemize}[nosep,leftmargin=*]
    \item \textbf{query\_local\_place}: Retrieves place information from cache
    \item \textbf{query\_local\_coordinates}: Retrieves coordinates with geocode fallback
    \item \textbf{query\_local\_routes}: Retrieves cached route summaries
    \item \textbf{query\_local\_travel\_time}: Retrieves cached travel time estimates
    \item \textbf{query\_local\_places\_batch}: Batch place information retrieval
    \item \textbf{query\_local\_nearby\_places}: Retrieves cached nearby search results
\end{itemize}

\section{Fine-tuning Details}
\label{app:fine-tuning}

\subsection{Stage 1: Supervised Fine-Tuning}
Given a training set $\mathcal{D}_{\text{SFT}} = \{(q_i, V_i)\}_{i=1}^{N}$ of question-concept pairs, we minimize the negative log-likelihood:
\begin{equation}
\mathcal{L}_{\text{SFT}} = -\sum_{i=1}^{N} \log p_\theta(V_i \mid q_i)
\end{equation}
where $\theta$ denotes the model parameters, and $V_i$ represents the annotated concepts with their types from $\mathcal{C}$ and functional roles from $\mathcal{R}$.

\subsection{Stage 2: Direct Preference Optimization}
For each question $q$, we construct preference pairs $(G^+, G^-)$ where $G^+$ satisfies all well-formedness constraints $\text{Valid}(G^+) = \bigwedge_{i=1}^{5} \mathcal{C}_i(G^+)$, while $G^-$ violates at least one constraint, i.e., $\exists i : \neg\mathcal{C}_i(G^-)$. The DPO objective is:
\begin{equation}
\mathcal{L}_{\text{DPO}} = -\mathbb{E}_{(q, G^+, G^-)}\left[\log \sigma\left(\beta \cdot r_\theta(q, G^+, G^-)\right)\right]
\end{equation}
where $r_\theta(q, G^+, G^-) = \log \frac{p_\theta(G^+ \mid q)}{p_{\text{ref}}(G^+ \mid q)} - \log \frac{p_\theta(G^- \mid q)}{p_{\text{ref}}(G^- \mid q)}$, $\sigma$ is the sigmoid function, $\beta$ is the temperature parameter, and $p_{\text{ref}}$ is the reference model (initialized from Stage 1). This formulation encourages the model to prefer graphs satisfying $\mathcal{C}_1$ (acyclicity), $\mathcal{C}_2$ (role ordering), $\mathcal{C}_3$ (type compatibility), $\mathcal{C}_4$ (data availability), and $\mathcal{C}_5$ (connectivity).

\section{Template Library}
\label{app:templates}

Our system employs a library of macro-templates that cover the major spatial reasoning patterns:

\paragraph{\textsc{Filter-Aggregate-Measure}} Filters objects by spatial/temporal predicates, aggregates results, and computes a final measure.

\paragraph{\textsc{Object-Field-Measure}} Transforms discrete locations into continuous fields (distance, bearing) for measurement.

\paragraph{\textsc{Route-Optimize}} Computes optimal visiting order for multi-stop trips with time windows.

\paragraph{\textsc{Geocode-Batch-Compare}} Batch geocodes candidates and compares against a reference point.

\paragraph{\textsc{Location-Bearing-Classify}} Computes bearing angle and converts to cardinal direction.

\paragraph{\textsc{Route-Step-Extract}} Analyzes navigation steps for specific maneuvers.

\paragraph{\textsc{Multi-Route-Compare}} Computes multiple candidate routes and selects by metric.

\paragraph{\textsc{Place-Attribute-Query}} Retrieves POI details and queries temporal/rating attributes.

\paragraph{\textsc{Multi-Segment-Aggregate}} Computes multi-leg journeys with mixed transport modes.

\paragraph{\textsc{Time-Window-Reverse}} Reverse-calculates latest departure given a deadline constraint.

\subsection{Example Retrieval Mechanism}
To provide semantic guidance during compositional generation, we maintain an example store $\mathcal{E} = \{(q_i, G_i)\}_{i=1}^{N}$ of question-graph pairs. Each graph $G_i$ is serialized into a textual description. Given an input question $q$, we retrieve the top-$k$ most similar examples by computing cosine similarity between the embedding of $q$ and each stored question $q_i$. The retrieved examples guide parameter binding and edge instantiation during template-based composition.

\section{Execution and Response Generation Details}
\label{app:execution}

\subsection{Execution Semantics}
Let $\Sigma: V \rightarrow \mathcal{D}$ denote the concept state, mapping each entity to its resolved data value in domain $\mathcal{D}$. Starting from an initial state $\Sigma_0$ (with extent nodes grounded to input data), we iteratively apply each operator to update the state until all nodes are resolved.

The execution trace $\mathcal{F} = [(s_1, \Sigma_1), (s_2, \Sigma_2), \ldots, (s_M, \Sigma_M)]$ records the state snapshot after each operator execution. For example, given a place name ``Central Park'', the first operator \texttt{geocode} resolves it to coordinates $(40.78, -73.97)$, updating $\Sigma_1$; a subsequent \texttt{haversine} operator then computes the distance to another location, yielding $\Sigma_2$ with the distance value. This trace enables interpretability and debugging.

\begin{algorithm}[t]
\caption{GeoFlow Graph Execution}
\label{alg:execution}
\begin{algorithmic}[1]
\Require GeoFlow Graph $G = (V, E, \lambda, \rho)$, initial state $\Sigma_0$
\Ensure Final state $\Sigma_M$, execution trace $\mathcal{F}$
\State $T \gets \text{TopologicalSort}(G)$ \Comment{Order by role priority}
\State $\Sigma \gets \Sigma_0$; $\mathcal{F} \gets []$
\For{each step $s_i = (B_i, A_i, \omega_i, \theta_i) \in T$}
    \State $\text{inputs} \gets \{\Sigma(v) : v \in B_i\}$
    \State $\text{outputs} \gets \omega_i(\text{inputs}; \theta_i)$
    \For{each $v_j \in A_i$}
        \State $\Sigma(v_j) \gets \text{outputs}[j]$
    \EndFor
    \State $\mathcal{F}.\text{append}((s_i, \Sigma))$
\EndFor
\State \Return $\Sigma$, $\mathcal{F}$
\end{algorithmic}
\end{algorithm}

\subsection{Grounded Response Generation}
The executed GeoFlow Graph yields a final state $\Sigma_M$ containing structured geospatial evidence (e.g., computed distances, optimized routes, filtered place rankings, and temporal constraints) derived through the agent's tool invocations and spatial computations. The agent generates its final response by grounding on these verifiable intermediate results produced during graph execution.

Formally, the agent synthesizes the final response by conditioning on both the original question and the execution outcomes:
\begin{equation}
a = \mathcal{L}_{\text{gen}}(q, \Sigma_M, \mathcal{F})
\end{equation}
where $\mathcal{F}$ denotes the execution trace providing step-by-step reasoning evidence. This formulation enables \textit{grounded generation}: the response is factually anchored in computational results obtained through tool-augmented reasoning, rather than hallucinated from learned priors.

\end{document}